\documentclass[11pt,a4paper]{article}
\usepackage[hyperref]{acl2021}
\usepackage{times}
\usepackage{latexsym}

\usepackage{xcolor,colortbl}
\usepackage{amsmath}
\usepackage{graphicx}
\usepackage{booktabs}
\usepackage{caption}
\usepackage{subcaption}

\usepackage{color}
\usepackage{multicol}
\usepackage{url}
\usepackage{multirow}
\usepackage{amssymb}
\usepackage{algorithm}
\usepackage{algpseudocode}

\usepackage{microtype}
\newcommand{\debjit}[1]{{\color{brown} #1}}
\newcommand{\af}[1]{{\color{blue} #1}}
\newcommand{\red}[1]{{\color{red}  #1}}
\newcommand{\orange}[1]{{\color{orange}  #1}}

\newcommand{\blue}[1]{{\color{blue} #1 }}

\newcommand{\bleu}{\textsc{Bleu}}
\newcommand{\COINS}{\textsc{Coins}}
\newcommand{\COMET}{\textsc{Comet}}
\newcommand{\ATOMIC}{\textsc{Atomic}}
\newcommand{\GLUCOSE}{\textsc{Glucose}}

\title{COINS: Dynamically Generating \underline{CO}ntextualized \underline{I}nference Rules for \underline{N}arrative \underline{S}tory Completion}

\aclfinalcopy
\author{Debjit Paul \\
  Research Training Group AIPHES \\
  Institute for Computational Linguistics\\
  Heidelberg University \\
  {\tt paul@cl.uni-heidelberg.de} \\\And
  Anette Frank \\
  Research Training Group AIPHES \\
  Institute for Computational Linguistics \\
  Heidelberg University\\
  {\tt frank@cl.uni-heidelberg.de} \\}

\date{}

\begin{document}
\maketitle
\begin{abstract}

Despite recent successes of large pre-trained language models in solving reasoning tasks, their inference capabilities remain opaque. We posit that such models can be made more interpretable by explicitly generating interim inference rules, and using them to guide the generation of task-specific textual outputs. In this paper we present \textsc{Coins}, a recursive inference framework that i) iteratively reads context sentences, ii) dynamically generates contextualized inference rules, encodes them, and iii) uses them to guide task-specific output generation. We apply \COINS\ to a \textit{Narrative Story Completion} task that asks a model to complete a story with missing sentences, to produce a coherent story with plausible logical connections, causal relationships, and temporal dependencies. By modularizing inference and sentence generation steps in a recurrent model, we aim to make reasoning steps and their effects on next sentence generation transparent. Our automatic and manual evaluations show that the model generates better story sentences than SOTA baselines, especially in terms of coherence. We further demonstrate improved performance over strong pre-trained LMs in generating commonsense inference rules. The recursive nature of \COINS\ holds the potential for controlled generation of longer sequences.

\end{abstract}
\section{Introduction}
\begin{figure}[t]
  \centering
    \includegraphics[scale=1.0,height=3.6cm, width=0.35\paperwidth]{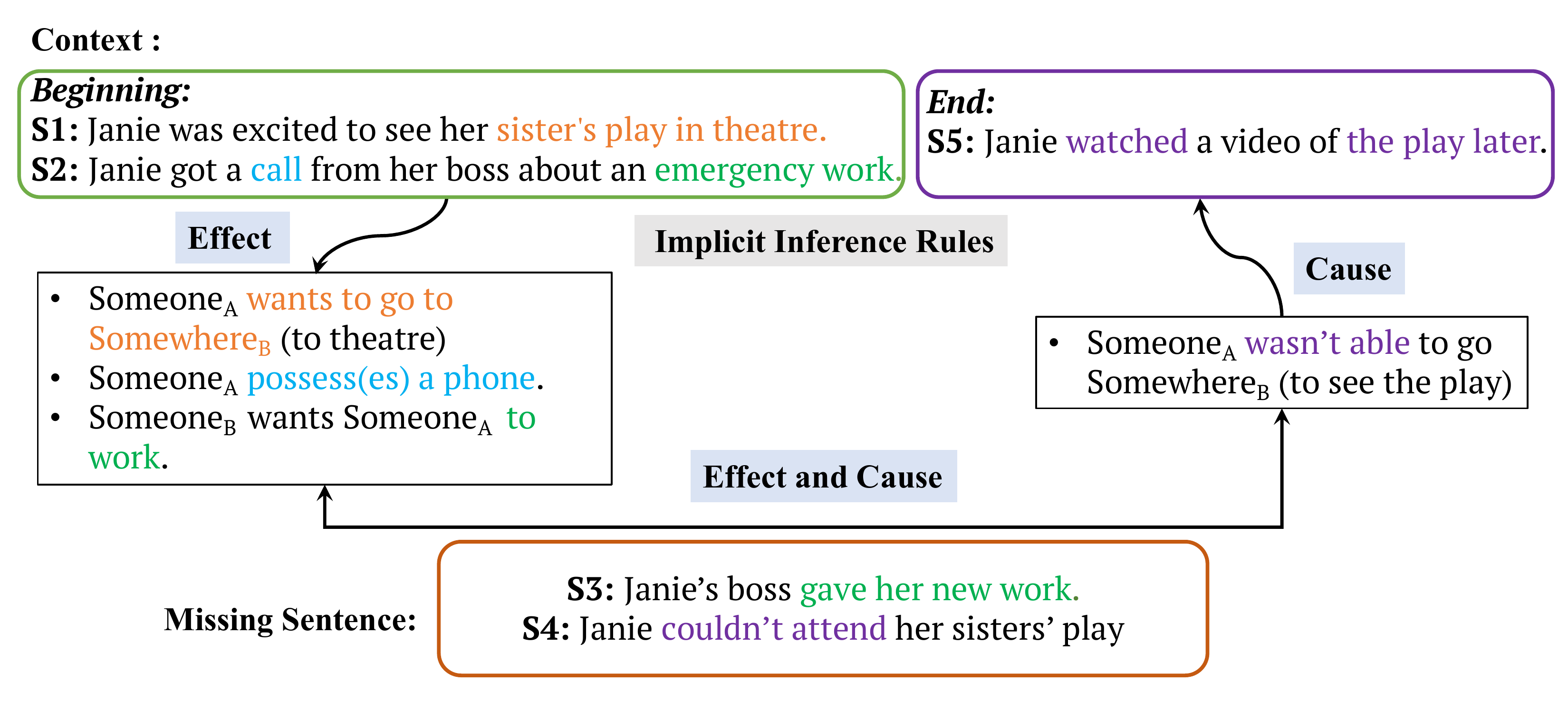}
    \caption{
    An example of the \textit{Narrative Story Completion Task}.
    Top and bottom boxes show the context (top) and missing sentences (bottom). The chain of implicit inference rules explains the
    connection between beginning and end, and allows to infer the missing sentences.
    }
    \label{fig:example_motivation}
\end{figure}

Narrative story understanding, and similarly story generation, requires the ability to construe meaning that is not
explicitly stated through commonsense reasoning over
events in the story \cite{rashkin2018modeling}. Previous work in modeling narrative stories
has focused on learning scripts\footnote{Scripts are structured knowledge about stereotypical event
sequences together with their participants.} \cite{welin1979scripts, mooney1985learning} and learning 
narrative schemas using
corpus statistics \cite{chambers-jurafsky-2009-unsupervised, balasubramanian-etal-2013-generating, nguyen2015generative}. 
Recently, large pretrained language models (LMs) such as GPT-2 have shown remarkable performance on various generation tasks. 
While these pretrained LMs
learn probabilistic associations
between words and sentences, they still have difficulties in modeling causality \cite{mostafazadeh-etal-2020-glucose}. 
Also, in narrative story generation, models need to be consistent with 
everyday commonsense norms. 
Hence, to address a story generation task, i) models need to be equipped with suitable knowledge, ii) they need
effective knowledge integration and reasoning methods, and ideally iii) we want to be able to make the effectiveness of these methods transparent. 

In this work we focus on the aspects i) to iii), 
by investigating new methods that build on
pretrained LMs to 
generate missing sentences from an incomplete narrative story. Specifically, we focus on \textit{Narrative Story Completion (NSC)}, a new task setting for story generation. Given an incomplete story, specified only through its beginning and ending, the task is to generate the missing sentences 
to complete the story (see Figure \ref{fig:example_motivation}).
Our hypothesis is that in order to obtaining a consistent and coherent narrative story, the task
%
 requires a model's ability to perform commonsense inference about 
 events and entities in a story. 
Unlike other
existing tasks, NSC
requires: \textit{i)} generating \textit{multiple sentences} to complete a story, and \textit{ii)} \textit{ensuring that} the generated sentences 
are \textit{coherent} with respect to both \textit{beginning and ending} of the story. Hence, the NSC task offers a challenging setup for investigating the reasoning capacities of a story generation
model. 

Humans excel in drawing inferences and constructing 
causal chains that explain the connection between events \cite{kintsch1978toward}. 
Figure \ref{fig:example_motivation} illustrates this with an example from our \textit{NSC}
task.\footnote{We use the ROCstories dataset to frame the \textit{NSC} task.}
From
 \textit{Janie was excited to see her sister's play in theatre}$_{(s_1)}$.  \textit{Janie got a call from her boss about new work}$_{(s_2)}$ and the outcome \textit{Janie watched a video of the play later.}$_{(s_5)}$ -- we can 
 construct inference rules 
 in forward and backward direction: forward via \textsc{Effect}:  Someone$_B$ (\textit{boss}) gave work to Someone$_A$ (\textit{Janie}); backward via \textsc{Cause}: Someone$_A$ (\textit{Janie}) wasn't able to go Somewhere$_B$ (\textit{to the theatre}). 
By combining these 
inferences, we can obtain a representation from which to generate a connection that completes the story, e.g., \textit{Janie's boss wanted her to look after the issue$_{(s_3)}$. She missed the theatre play$_{(s_4)}$}.


In this work, we propose \textsc{Coins}: a recursive model that jointly learns to i) \textit{dynamically} generate commonsense inference rules\footnote{In this paper, similar to \citet{mostafazadeh-etal-2020-glucose}, we will use “inference rule” and “explanation” interchangeably.} grounded in the context and  to ii) perform \textit{controled and coherent}
story generation, using the generated inferences as a guide.
We hypothesize that jointly learning to generate contextualized inference rules from
\textit{dynamically predicted contextualized inference rules}
and learning to generate story sentences \textit{incrementally} while taking the inferences into account, will 
improve the quality of \textit{both} the predicted inference rules and of generated story sentences.
Moreover, the recursive nature of the model and the \textit{individuation}
of the inference prediction 
and sentence generation tasks
make the process 
more \textit{interpretable}:
the generated inference rules can be viewed as intermediate representations,  and 
can serve as 
\textit{explanations} of how the 
dynamically produced inferences 
influence the quality of 
generated story sentences.

Our main contributions 
are
as follows:  

1) We propose a new setting for a Narrative Story Completion task, which asks
a system to complete a narrative story given its beginning and ending, 
with the aim of examining 
the reasoning capacities of a model that solves the task.

2) We propose an integrated reasoning and NL generation model, \COINS, that based on its cur\-rent context generates contextualized commonsense inference rules and follow-up 
sentences, in a step-wise recurrent pro\-cess.

3)  We conduct extensive experiments with automatic and human evaluation.
Automatic evaluations show that \COINS\ outperforms 
strong baselines ($+2.2$ \bleu\ score). Human evaluation shows that compared to strong
baselines, our model yields better sentence generations
with respect to \textit{coherence} ($+50.5 \%$) and \textit{grammaticality} ($+20.5 \%$).


4)  We 
show that \textsc{Coins} 
generates better 
inference rules ($+2.3$ \bleu\ score) compared to a \textit{fine-tuned} GPT-2 
model, and that jointly learning to generate inferences and story sentences improves the quality of the generated inference rules.

Our code is made publicly available.\footnote{\url{https://github.com/Heidelberg-NLP/COINS}}

 

\if false 
Human understanding of narrative texts requires making commonsense inferences beyond what is stated explicitly in the text ......
Humans make countless implicit commonsense inferences about everyday situations .....
Humans can construct the causal chain that describes the sequence of events led to a particular outcome. 
Humans make countless implicit commonsense inference that frames there understanding of the unfolding narrative. 
\fi 
\section{Related Work}
\textbf{Sentence-level Commonsense Inference and Beyond.} 
Recent research in this area has
focused on commonsense knowledge acquisition \cite{sapatomic, zhang2020aser, speer2017conceptnet, malaviya2019exploiting} and commonsense reasoning \cite{zellers2019hellaswag,Talmor2018CommonsenseQAAQ}. In our work, we focus on inferential knowledge about events, and entities participating in such events. \citet{rashkin-etal-2018-event2mind} introduced
a knowledge resource 
of
commonsense inferences regarding
people's intents and reactions towards
a diverse set of events.
With \COMET, \citet{bosselut-etal-2019-comet} 
have shown that pre-trained neural language models can be fine-tuned using large knowledge bases (such as \ATOMIC, \citet{sapatomic}) to generate 
inferences for a given event or sentence.
However, the generated knowledge from \COMET\ is non-contextualized and hence, can be inconsistent. Recently, \citet{mostafazadeh-etal-2020-glucose} proposed \textsc{Glucose}, a new resource and dataset that offers semi-structured commonsense inference rules that are \textit{grounded} in sentences of specific stories. They show that fine-tuning a pre-trained LM
on the \textsc{Glucose} dataset helps the model to better generate inferrable commonsense explanations given a \textit{complete} 
story. In concurrent work, \citet{Gabriel2021ParagraphLevelCT} proposed PARA-COMET, a model that incorporates paragraph-level information to generate coherent commonsense inferences from narratives.
In this work, we investigate how well a neural model can generate contextualized commonsense inference rules for an \textit{incomplete} 
story. 
Learning to predict iterative inference steps for successive events in a narration using
semi-structured knowledge rules 
is still a difficult and underexplored task. We propose a model that learns to iteratively generate a coherent completion of an incomplete narrative story utilizing semi-structured knowledge as 
offered
by the \textsc{Glucose} framework.



\textbf{Commonsense Reasoning in
Narrative Stories.} Early work on 
narrative events focused on \textit{script learning}, by defining stereotypical event sequences together with their participants \cite{welin1979scripts}.
In later works, \citet{chambers-jurafsky-2008-unsupervised, chambers-jurafsky-2009-unsupervised, balasubramanian-etal-2013-generating, nguyen2015generative, pichotta2014statistical} proposed methods to learn 
\textit{narrative event chains} using a simpler event representation that allows for efficient learning and inference.
\citet{chambers-jurafsky-2009-unsupervised} acquired Narrative Event Schemata from corpora and established the Narrative Cloze Task \cite{chambers-jurafsky-2008-unsupervised} that evaluates script knowledge by predicting a missing event (verb and its arguments) in a sequence of observed events.
More recently, \citet{mostafazadeh-etal-2016-corpus} proposed the \textit{story cloze task} that selects a plausible (right) over an implausible (wrong) story ending.
\citet{bhagavatula2019abductive} proposed an \textit{abductive reasoning task} to test a model’s ability to generate plausible explanations for
an incomplete set of observations. \citet{paul-frank-2020-social} proposed a multi-head knowledge attention method to dynamically incorporate non-contextualized inferential knowledge to address the \textit{abductive reasoning task}. \citet{qin-etal-2020-back} proposed an unsupervised decoding algorithm that can flexibly incorporate both the past and future contexts using only off-the-shelf language models to generate plausible explanations. Concurrent to our work, \citet{paul-frank-2021-generate-hypothetical-events} presented a method for addressing the \textit{abductive reasoning task} by explicitly learning what events could follow other events in a hypothetical scenario. 
In our work, we make use of the ROCStories dataset \cite{mostafazadeh-etal-2016-corpus} to build a \textit{Narrative Story Completion} task that tests a model's ability of \textit{generating} missing sentences in a story. We propose a model that aims to produce
\textit{coherent} narrative stories by performing iterative commonsense inference steps.



\textbf{Narrative Story Generation.} Much existing work on story generation
relied on symbolic planning methods \cite{lebowitz1987planning, doi:10.1080/09528130010029820, Jzefowicz2016ExploringTL}. With the advances of Seq2Seq models, several works 
applied them in automatic story generation tasks \cite{roemmele2016writing, Jain2017StoryGF}. \citet{fan-etal-2018-hierarchical} proposed a hierarchical approach to 
generate short stories from initial prompts. 
Recently, many works have focused on integrating external commonsense knowledge from large static knowledge bases like \ATOMIC\ \cite{sapatomic} or ConceptNet \cite{speer2017conceptnet} for different tasks
such as story ending generation 
\cite{ji-etal-2020-language, Guan_Wang_Huang_2019} or story generation 
\cite{guan-etal-2020-knowledge, xu-etal-2020-megatron}. 
In
concurrent work, \citet{ammanabrolu2020automated} look into causality for a commonsense plot generation task. In our work, we model the assumption that contextualized inference rules provide \textit{inferred} information that can guide a system in generating both \textit{contextually grounded} and \textit{coherent} follow-up
sentences in a story generation task.

\if false  

by multi-source attention mechanism \cite{guan2019story}

\textbf{Reasoning about Narratives.} A prominent resource from recent years is the RocStories corpus
(Mostafazadeh et al., 2016b), consisting of 98K
crowdsourced 5-sentence everyday life stories. It
was used for the story cloze task whose goal was to
predict the story ending from its first 4 sentences,
but gained popularity and became the base of additional benchmarks (Rashkin et al., 2018). Additional related work includes “script knowledge”, i.e.
learning about prototypical series of events (Schank
and Abelson, 1977; Chambers and Jurafsky, 2008;
Pichotta and Mooney, 2014), temporal commonsense (Granroth-Wilding and Clark, 2016; Li et al.,
2018), and modeling pre- and post- conditions of
events (Roemmele et al., 2011; Sap et al., 2019;
Bosselut et al., 2019). Qin et al. (2019b) studied
conversation modeling that reads and connects the
dots of events in related documents. Finally, a recent line of work explores counterfactual questions
in reading comprehension (Huang et al., 2019; Tandon et al., 2019), but instantiates the problem of
counterfactual reasoning as a multiple choice task.

Several prior work ... 

Automatic story generation is a longstanding problem in AI, with early work dating back to the 1970s
based on symbolic planning (Lebowitz, 1987;
Perez y P ´ erez and Sharples ´ , 2001; Porteous and
Cavazza, 2009; Riedl and Young, 2010) and casebased reasoning using ontologies (Gervas et al. ´ ,
2005). Li et al. (2013) extended prior works toward learning domain models (via corpus and/or
crowdsourcing) to support open story generation
about any topic.
\\

\fi 
\section{Task Definition}
We formulate the \textit{Narrative\ Story\ Completion task (NSC)} as follows: given an incomplete story ($S$= $s_1, s_2, s_n$) as a sequence of tokens $t$ = $\{t_1, t_2, ..., t_{SEP}, ..., t_m\}$ (with $t_{SEP}$ a mask token delimiting $s_2$ and $s_n$), the goal 
is to generate the missing sentences ($s_3, ... , s_{n-1}$) as a sequence of tokens $y^{s_i}$=\{$y^{s_i}_1, y^{s_i}_2, ... ,y^{s_{i}}_v$\} (with $i = 3$, ..., $n$$-$$1$ and
$v$ 
the maximum length of each sentence). 

In the setting of the NSC task, we expect 
the completed story to be coherent. That is, the generated sentences should exhibit reasonable logical connections, causal relationships, and temporal dependencies with each other and 
the given beginning and ending of the story. 
In this paper, we define a discourse to be
coherent if successive sentences that are about the same entities, and the reported events 
involving them can be construed to reflect common knowledge about how events are typically connected in a temporal sequence or by causal relations. Similar to \citet{hobbs1985coherence}, the criteria to conclude that discourse is coherent include require that
there are reflections of causality in the text. 

Our take on this task is to incrementally generate contextualized inference rules from the given context, and to make use of this knowledge to generate missing story sentences. 

\section{Discourse-Aware 
Inference Rules}
\begin{table}[!tbp]
      \scalebox{0.7}{
      \centering
      \begin{tabular}{@{}p{23mm}|p{80mm}}
          \hline
            \textbf{Relation Type} & \textbf{Dimensions} \\  \hline
            {\textbf{Cause}\linebreak (Dim 1-5)} & {
            (1) Event that directly causes or enables X; (2) Emotion or basic human drive that motivates X;  (3) Location state that enables X;  (4) A possession state that enables X; (5) Other attribute that enables X.} \\
            {\textbf{Effect}\linebreak (Dim 6-10)} & {
            (6) An event that is directly caused or
enabled by X; (7) An emotion that is caused by X; (8) A change of location that X results in; (9) A change of possession that X results in; (10) Other change in attribute that X results in.} \\\hline
      \end{tabular}}
      \caption{Causal Relation types and their mapped relations \cite{mostafazadeh-etal-2020-glucose}.}
      \label{tab:glucose}
\end{table}

This section details 
how we construct
training data
for the NSC task, by enriching stories with automatically predicted contextualized inferences.\footnote{For testing  we rely on \GLUCOSE's manually validated inference rules on a small subset of the  ROCStories corpus.} We utilize the \GLUCOSE\ \cite{mostafazadeh-etal-2020-glucose} dataset, which contains implicit commonsense knowledge in form of semi-structured general and specific 
inference rules\footnote{\textit{Specific} means rules
grounded in a given context and \textit{general} corresponds to rules that are applicable to other contexts.} (cf. Table \ref{tab:glucose})
that are grounded in the context of individual stories from ROCStories.
In \GLUCOSE, given a story $S$ and a selected sentence $X$ from the story, the authors define ten dimensions $d$ of commonsense causal explanations related to $X$, inspired by human cognitive psychology. Only a small part of  ROCStories is annotated with \GLUCOSE\ inferences (Table \ref{tab:data_stat}). 


Given the amount of commonsense knowledge needed for real-world tasks, 
a static
knowledge resource is always incomplete.
Thus, we
\textit{fine-tune} a pre-trained GPT-2 model on the annotated part of \textsc{Glucose} to 
\textit{dynamically} generate inference rules for each sentence $X_i$ of each story $S_i$ from the underlying
ROCStories data. We \textit{fine-tune} two separate language models $CSI_{gen}$ and $CSI_{spec}$ for general and specific rules, respectively (Table \ref{tab:inferenceexamples}).  
\if false 
Contrary to the training input sequence ($S\#X\#d\#answer\#EOS$) in \GLUCOSE, we 
\red{[I don't think that you need the concrete input format here. what matters here is that in \textsc{Glucose} for each X, a system predicts/generates  inferences for each dimension $d$ separately, while you cluster the different dimensions into two clusters/classes (EFFECT and CAUSE), and make predictions/generate rules for the each of these aggregate classes. It does not matter what 'paraphrases' or 'full rules' are, at this point. Or does it? Can you rather write a footnote that says: we ignore paraphrases used in \GLUCOSE?]}
where $d$ is the dimension number and $answer$ is the full rules\footnote{\debjit{\textit{full rules} includes the contextualized rules, dimension and the paraphrase of $X$.}}, 
in our setting we cluster all dimensions $d$ into two categories \textsc{Effect} vs.\ \textsc{Cause} (cf. Table \ref{tab:glucose}). Hence, our input sequence ($S\#X\#r\#answer\#EOS$) where $r$ is either \textsc{Effect} or \textsc{Cause}. We concatenate all the rules across the dimension for each relation to create $answer$ tokens. Details regarding the input formats can be found in Appendix. \red{[I'd much prefer to have a simple format here, that is explained and does not ask me to go to the APPENDIX.]}

\red{please START OUT by explaining why you want to aggregate the predictions, rather than asking the reader to follow your definitions, and only by the end get to know why this is good]}
The intuitive reason to concatenate all the rules is that we want the model to learn both: (i) which dimensions can be inferred for a given incomplete story ($S'$), sentence ($X$) and a relation ($r$), and (ii) the full rules for those dimensions. \red{[I did still not understand what 'full' rules are. Is it needed?]} For example, in Fig. \ref{fig:example_motivation}, for $s_2$ it can be inferred that Janie \textit{possess} a phone, whereas for $s_5$ the dimension \textit{possession} doesn't hold.  \red{[This is not well explained: also when aggregating, the system needs to decide whether to generate a possession-related CAUSE or not. In discussion you raised an efficiency issue.]}
\fi 
\begin{table}[t]
\centering
\small
\scalebox{0.8}{
\begin{tabular}{@{}p{12mm}|p{78mm}}
\rowcolor{blue!15}
Incomplete Story: & $s_1$: Jane loved cooking. $s_2$: Everyone else in her family did too.
$s_5$: \textbf{Eventually she learned everything there was to teach.} \\
\textit{Gold}: & Someone$_A$ loves  Something$_A$ (that is an activity ) $>$\textsc{Causes/Enables}$>$ Someone$_A$ learns everything there is to learn. \\
\rowcolor{gray!15}
&{Jane  loves  cooking  $>$\textsc{Causes/Enables}$>$ Jane learns everything there is to learn} \\
\textit{COINS}: & Someone$_A$ is a quick learner $>$\textsc{Causes/Enables}$>$ Someone$_A$ learns everything there is to learn. \\
\rowcolor{gray!15}
 & {Jane is a quick learner $>$\textsc{Causes/Enables}$>$ Jane learns everything there is to learn.} 
\end{tabular}}

\caption{Example of inference rules generated by \COINS\ (compared to \textit{Gold} from \GLUCOSE). Grey: 
context-specific rules (SR); regular: 
general rules (GR). 
Bolded sentence $s_5$ is X, \textsc{Cause} 
is the relation type $r$.}
\label{tab:inferenceexamples}
\end{table}

\begin{table}[t!]
\small

\centering
{
\scalebox{0.9}{
\begin{tabular}{@{}lllll@{}}
\toprule
{\bf Dataset} & \bf{Relation Type} &{\bf Train}& {\bf Dev} & {\bf Test}\\
\midrule
\textbf{NSC} &&{88,344}& {4,908} & {4,909} \\
\textbf{GLUCOSE} & \bf Effect & {2949} & {849} & {--}\\
\textbf{} & \bf Cause & {2944} & {916} & {--}\\
\bottomrule
\end{tabular}}}
\caption{Dataset Statistics: number
of unique stories.}\label{tab:data_stat}
\end{table} 
The 10 dimensions $d$ in \textsc{Glucose} cover implicit \textit{causes} and \textit{effects} of a sentence $X$ in a given story. In our work, we are interested in inference rules that explain a sentence's causes and effects, to study the impact of such inferences
on narrative story completion. We therefore cluster all dimensions $d$ into the two categories \textsc{Effect} vs.\ \textsc{Cause} (Table \ref{tab:glucose}) and aggregate 
all rules from the respective categories 
(preserving their dimensions). 
Once our models ($CSI_{gen}$, $CSI_{spec}$) are trained, we apply them to our \textit{NSC} task training data,
to enrich it with  inference rules for each sentence and story.


\section{\COINS: COntextualized Inference and Narrative Story Completion Model}

\begin{figure}[t]
  \centering
    \includegraphics[scale=1.0,height=5cm, width=0.47\textwidth]{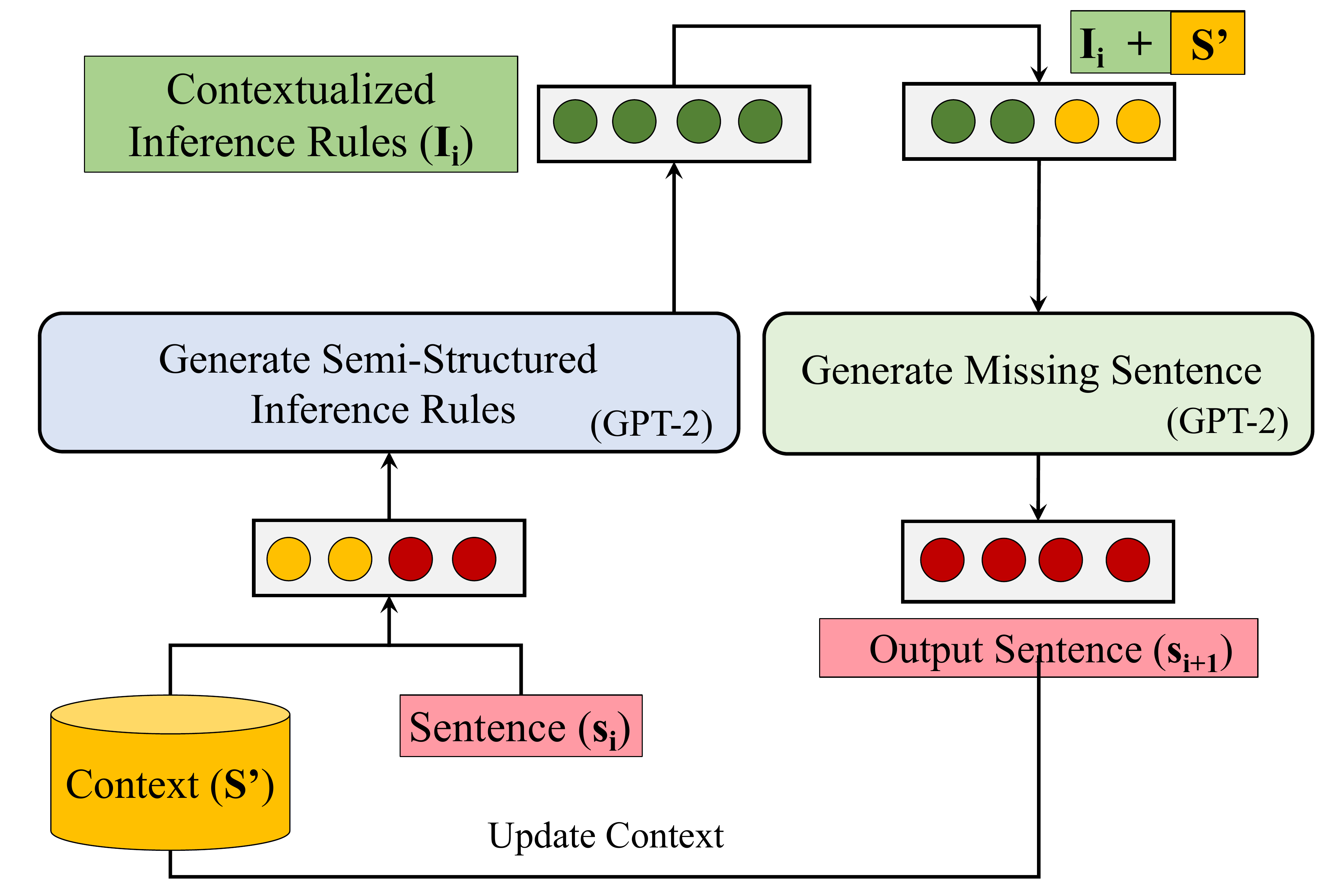}
    \caption{Architecture of the
    COINS model.}
    \label{fig:coins}
\end{figure}

In this section we introduce a recursively operating reasoning and sentence generation model: \COINS. An overview 
is given in Figure \ref{fig:coins}. In each iteration, the model applies two consecutive steps:\\
(1) \textit{Inference Step}: Given an \textit{incomplete story context $S'$}= $X \oplus S_i$ and   
relation $r$, 
an \textit{inference model} $CSI$ ($gen$ or $spec$) generates COntextualized inference rules of type $r$.\\ 
(2) \textit{Generation Step}: a \textit{sentence generator} reads
the generated inference rules concatenated with the
current context $S'$ and generates the next 
story sentence $s_{i+1}$. The context $S'$ is updated with $s_{i+1}$
and steps (1) and (2)
are repeated (cf.\ Algorithm 1).\\

This formulation allows us to i) examine inference and generation capabilities separately from each other, ii) helps determine the impact of inferential knowledge 
on story generation, 
and iii) can give us insight  into  how  knowledge  can  guide  story  
generation in a recursive inference framework.

\paragraph{Inference Step.} 
We define
the initial story context $S'$ = $\{s_1, s_2,$[SEP]$, s_n\}$, a selected sentence as $s_i$, and relation type $r \in$ \{\textsc{Effect}, \textsc{Cause}\}, where $i \in [2,\dots $$n$-$1]$, $s_i$=\{${w^{s_i}_1,..,w^{s_i}_v}$\}.  
We adopt a pretrained GPT-2 (base) \cite{radford2019language} transformer model with
multiple Transformer blocks of multi-head self-attention and fully connected layers.
During training, in each iteration the input to the model is a concatenation of the current source ($S',s_i,r$) and target sequence i.e., the inference rules ($E_i$ or $C_i$). 
Eq.\ (1) defines the inference rule ($\mathcal{IR}$) generation model:

\vspace*{-3mm}
\begin{align}
\label{eqn:gpt2}
\begin{split}
 h^0_p = e_p+P_p,
\\
 h{_p^l} = block(h^{l-1}_{<p}), l \in [1,L] 
 \\
 p(y_p|y_{<p},p) = softmax(h{_p^L}W^T)
\end{split}
\end{align}
where $h^0_p$ is a summation of token embedding $e_p$ and position embedding $P_p$ for the $p$-th token; $h_p^l$ is the $l$-th layer's output at position $p$,
computed through transformer blocks with the masked multi-head self attention mechanism; $h{_p^L}$ is the final layer's hidden state and $y_{<p}$ indicates the left context of position $p$. The softmax layer defines the model to output the most probable target sequence: the most likely 
inference rules ($E_i$ and $C_i$) for each relation  type (cf. Algorithm Line 4-5). 

During training, we minimize the objective (2) \vspace*{-3mm}

\begin{align}
\label{eqn:lossinferences}
\begin{split}
 \mathcal{L}_{I}(\beta)= -\sum_{k=m}^{m+N}{log\ p(E_{i}^{k}| S', s_{i}, \textsc{Effect})} \\ -\sum_{k=m}^{m+N}{log\ p(C_{i}^{k}| S', s_{n}, \textsc{Cause})}
\end{split}
\end{align}
where $m, N$ denote the number of tokens in the source 
($S',s_i,r$) and target sequence (inference rules) respectively; $\beta$ refers to model parameters. 

In this work, we focus on the NSC
task, which requires our model to capture temporal dependencies and causal relationships between events. While we 
designed 
our sentence generation model in such a way that it
can 
utilize inference rules from both forward and backward directions for each sentence, we here
trigger the generation of \textsc{Cause} inference rules for $s_n$, since we expect that \textit{events},  \textit{motivations} or \textit{attributes} that \textbf{cause} $s_n$ will be relevant for generating the preceding sentences $[s_3, \dots s_{n-1}]$.
Similarly, we generate \textsc{Effect} relations for $s_i$, assuming that 
an \textit{event}, changes of \textit{emotion} or changes of \textit{attribute} that are possible \textbf{effects}   
caused by $s_i$ will be most relevant for generating the missing follow-up sentences.
In principle, however, for NSC and other story generation tasks, we may consider \textsc{Cause} and \textsc{Effect} relations for all sentences, letting the model freely choose from the full space of inferences.


We concatenate the generated inference rules ($I_i=E_i \oplus C_i$)\footnote{We use $[SEP]$ token to delimit the individual E$_i$ and C$_i$ when concatenating them.} and store the last hidden representation in $Mem_\mathcal{IR} \in {\rm I\!R}^{N\times L \times H}$, where $N$ is the number of sentences, $L$ the maximum inference sequence length and $H$ the hidden state dimensions. 
$Mem_\mathcal{IR}$ is updated with
the hidden representations of inference rules in
each iteration. Hence, $Mem_\mathcal{IR}$ could act 
as an intermediate representation, and as a basis for providing \textit{explanations} for observed story sentence generations.
$Mem_\mathcal{IR}$ may also be used as a memory for long-form text generation tasks, to keep track of implicit knowledge \textit{triggered by}
previously generated text, and could support flexible discourse serialization patterns.\footnote{We leave such extensions to future work.}
\begin{algorithm}[t] 
\small
\caption{\textsc{Coins}} \label{generate}
\begin{algorithmic}[1]
\Require{Initial Context ($S'$ = \{$s_1, s_2,[SEP], s_n$\})}
\State $Mem_\mathcal{IR}$  $\gets$ empty 
\State $\mathcal{G}en\mathcal{S}$ $\gets$ empty list
\For{$i \gets 2$ to $n-1$}
    \State $E_{i}$ = \textit{GenInferenceRules}($S',s_{i}$, \textsc{Effect})
    \State $C_{i}$ = \textit{GenInferenceRules}($S',s_{n}$, \textsc{cause})
    \State $I_{i}$ = $E_{i}$ $\oplus$ $C_{i}$
    \State $s_{i+1}$ = \textit{GenNewSentence}${(I_{i}, S')}$ 
    \State $\mathcal{G}en\mathcal{S}$ := $\mathcal{G}en\mathcal{S}$ + $s_{i+1}$
    \State $Mem_\mathcal{IR}$ := $Mem_\mathcal{IR}$ $\oplus$ $I_{i}$
    \State $\mathcal{L_{S}}$ += $-log_{p_{(\theta)}}(s_{i+1}|I_{i},S')$  $-log_{p_{(\beta)}}(I_{i}|S')$
    \State $\mathcal{L_{IR}}$ += $-log_{p_{(\theta)}}(s_{i+1}|I_{i},S')$  $-log_{p_{(\beta)}}(I_{i}|S')$
    
    \State $S' := \{s_1, s_2, s_{i+1}, [SEP], s_n$\} 
\EndFor    
\State \Return $\mathcal{G}en\mathcal{S}$, $Mem_\mathcal{IR}$

\end{algorithmic}
\end{algorithm}
\paragraph{Generation Step.}

Given the generated inference rules $I_i$ (in form of  tokens) and the incomplete story context $S'$, 
we aim to generate the next missing sentence. We 
pass the input through another pretrained GPT-2 (base)
model (cf. Equation \ref{eqn:gpt2}). 
The loss function for the sentence generator is
\begin{equation}
    \mathcal{L}_{S}(\theta)= -\sum_{k=1}^v{log\ P(y^{s_{i+1}}_k|I_{i}, [EOK], S')}\label{eq:train}
\end{equation}
where $y_k$ denotes the $k$-th token 
and $v$ the maximum length of the generated
sentence; ${i \in [2,n-1]}$
; $[EOK]$ 
denotes the end of knowledge rule tokens, and $\theta$ refers to model parameters. 

\textbf{Update Story Context.} In the final step we update the story context by inserting the generated sentence $s_{i+1}$ into the previous story context (cf. Algorithm 1, line 12). 


\paragraph{Training and Inference.} 
We add the losses $\mathcal{L_I}$ for inference generation 
and $\mathcal{L_S}$ for sentence generation 
to make the models dependent on each other (Algorithm 1, line.\ 10-11). 
For both the inference and the generation step model,
we minimize the negative log likelihood loss of the 
respective target sequence.



\section{Experiments}
\subsection{Dataset}

We apply \COINS\
to  the \textit{NSC} and the \textit{Story Ending Generation} tasks.\footnote{The results for \textit{Story Ending Generation} will corroborate our results for \textit{NSC}. All details are given in the \textit{Appendix}.} For data statistics see Table \ref{tab:data_stat}. 
\textbf{Narrative Story Completion.} We follow the task definition as introduced 
in \S 3.\\ 
\textit{Data Collection.} We construct the \textit{NSC} dataset on the basis of 
the ROCStories corpus \cite{mostafazadeh-etal-2016-corpus}, which
contains 98,162 five-sentence stories 
with a clear beginning and ending, thus making it a good choice for this task. We choose the first two sentences ($s_1, s_2$) as beginning rather than just $s_1$ because the first sentence ($s_1$) tends to be short in length, and usually introduces characters or sets the scene \cite{mostafazadeh-etal-2016-corpus}, wherease the second sentence ($s_2$) provides more information about the initial story.

\subsection{Hyperparameter Details}

\textit{Parameter size.} For GPT-2 we use the GPT-2 small checkpoint (117M parameters) based on
the implementation of HuggingFace \cite{wolf-etal-2020-transformers}. \\
\textit{Decoding Strategy.} In the inference stage, we adopt beam search decoding with a beam size of 5 for all our models and all baselines we produce. \\
We used the following set of hyperparameters for our \textsc{Coins} model: batch size: \{$2, 4$\}; epochs: \{$3, 5$\}; learning rate: \{$1e$-$5$, $5e$-$6$\}. We use Adam Optimizer, and dropout rate = $0.1$. We ran our experiments with
GPU sizes of $11$GB and $24$GB.
\subsection{Baselines}
We compare our \COINS\ model 
to the following baselines:

(a) \textbf{GPT-2} \cite{radford2018improving} 
(with 12-layer, 768-hidden, 12-heads), trained with an objective to predict the next word.
The input to the GPT-2 model is the concatenation of the source and the target story sequence. We follow the standard procedure to fine-tune GPT-2 
on the NSC task during training and minimize the loss function: 
\begin{align}
\label{eqn:lossinferences}
\begin{split}
 -log(s_3, s_4 | [SOS] s_1, s_2, [SEP], s_5 [EOS])
\end{split}
\end{align}

(b) \textbf{Knowledge-Enhanced GPT-2} (\textbf{KE}) \citep{guan-etal-2020-knowledge} is the current SOTA for ROCStories generation. It first fine-tunes a pre-trained GPT-2 (small) model with knowledge triples from commonsense datasets (ConceptNet [CN] \citet{speer2017conceptnet} and \ATOMIC\ [AT] \citet{sap-etal-2020-commonsense}). The knowledge triples were converted to sentences using templates. A multitask learning framework further
fine-tunes this
model on both the \textit{Story Ending Generation task} and classifying corrupted stories from real ones. As our baseline we choose 
the version without multi-tasking, since the corrupted story setting is not applicable for the \textit{NSC} task.  

(c) \textbf{GRF} \cite{ji-etal-2020-language} is the current SOTA for the \textit{Abductive Reasoning} and the \textit{Story Ending Generation} tasks. GRF enables pre-trained models (GPT-2 small)
with dynamic multi-hop reasoning on multi-relational paths extracted from the external ConceptNet
commonsense knowledge graph. 

(d) \textbf{\GLUCOSE-GPT-2} Similar to \citet{guan-etal-2020-knowledge}, we \textit{fine-tune} pretrained GPT-2 (small) on the \GLUCOSE\ dataset using \textit{general rules} (GR). We follow the same procedure as \citet{guan-etal-2020-knowledge} and (i) first fine-tune a pre-trained GPT-2 , but here on the GLUCOSE dataset, with the following loss: 
\begin{align}
\label{eqn:lossinferences}
\begin{split}
 -log(I_i| S, s_i, r),
\end{split}
\end{align}
where r: \textsc{Cause/Effect}, $I_i$: Inference rules.
(ii) Then we fine-tune the above model again on the NSC dataset with the following loss:
\begin{align}
\label{eqn:lossinferences}
\begin{split}
 -log(s_3, s_4 | [SOS] s_1, s_2, [SEP], s_5 [EOS])
\end{split}
\end{align}

The main difference between GLUCOSE-GPT-2 and COINS is: \textbf{COINS} explicitly learns to generate (contextualized) inference rules \textit{on the fly}
during the inference step and incorporates them in 
the story generation step.

\subsection{Automatic Evaluation Metric}
For automatic evaluation in the \textit{NSC} task we use as metrics Perplexity (indicates fluency of text generation), \bleu-1/2 \cite{papineni2002bleu} and \textsc{Rouge-L} \cite{lin-2004-rouge}. We report performance on the test sets by
averaging results obtained for $5$ different seeds.
All improvements across all model variants are statistically significant at $p < 0.05$. 


\section{Results}
\begin{table}[t!]
\centering
{
\scalebox{0.68}{
\begin{tabular}{@{}l@{}c@{~~}c@{~~}c@{~~}c@{}}
\toprule
{\bf Model} & \textbf{PPL} ($\downarrow$) & \textbf{BLEU-1/2} ($\uparrow$) & \textbf{ROUGE-L} ($\uparrow$)\\\midrule

{\bf GPT-2} & 11.56 & 16.66/6.8 & 17.2\\
\textbf{KE} [CN, AT] & 12.61 & 17.55/7.6 & 17.9\\
\bf{GLUCOSE-GPT-2} & {12.7} & 17.9/7.8 & 17.5\\
\textbf{GRF} [CN] & {12.18} & 20.8/8.2 & 17.6\\
{\bf COINS (SR)} &\textbf{6.7}&  22.53/10.10 & 18.9 \\
{\bf COINS (GR)} &6.9& \textbf{22.82/10.52} & \textbf{19.4}\\\hline 
{\bf COINS Oracle (SR)} (Test-only) & --& {30.75/22.76} & {32.5}\\
{\bf COINS Oracle (GR)} (Test-only) & -- & {26.37/17.01} & {27.38} \\
{\bf{Human}} & -- & 24.53/12.10 & 20.2\\
\bottomrule
\end{tabular}}}
\caption{Automatic evaluation results for Story Completion. Best performance  highlighted in \textbf{bold}; used Inference Rule types: specific (SR), general (GR).
}\label{tab:automaticeval}\label{tab:resultnsc}
\end{table}

Our experimental results are summarised in Tables \ref{tab:resultnsc} and \ref{tab:automaticevalknowledge}. \\
\textbf{NSC task.} Table \ref{tab:resultnsc} shows the results for the models described in $\S$6.3 and evaluated as per $\S$6.4. We observe the following: (i) 
\COINS\ outperforms all strong baseline models that utilize pre-trained language models and incorporate external commonsense knowledge with respect to all automatic evaluation metrics.
Note that \textbf{\GLUCOSE-GPT2} and \textbf{\COINS} are using the same knowledge resource, hence the clear performance increase of \COINS\ ($+4.92$ \bleu\ score)
indicates that jointly learning to generate contextualized inferences rules and 
missing sentences in a recursive manner
can enhance generation quality.\footnote{Since \textbf{GRF}'s architecture is specific for ConceptNet, we cannot exclude that the better performance of \COINS\ ($+2.2$ \bleu) is in part due to differences in the used knowledge.} 
(ii) Similar to \citet{ji-etal-2020-language}  we observe that fine-tuning GPT-2
over knowledge triples ([\textsc{Cn}], [\textsc{At}]\textsc{omic} or [\textsc{Gl}]\textsc{ucose}) doesn't improve the overall performance by much (Table \ref{tab:resultnsc}, line 2: [\textsc{Cn+At}] vs.\ line 3: [\textsc{Gl}] vs.\ line 1: [no CSK]).
(iii) For \COINS, \textit{general rules} (GR) boost performance more than specific rules, indicating that the sentence generation model 
generalizes well. (iv) In the oracle settings at
inference time we provide the model with the silver inference rules (generated as per $\S$4) that use the complete story context as background. The result indicates that SR performs better than GR when the model sees the full story context. 

In general we observe that story generation benefits from higher-quality, contextualized
inference rules from \GLUCOSE\ (for \COINS).\footnote{Automatic (silver) \GLUCOSE\ inference rules (cf. \S4) of type GR yield $60.8$ \bleu\ score i.e., performance of $CSI_{gen}$ (avg.\ of both relation types).} 
The improvement of \COINS\ over \GLUCOSE-GPT-2 indicates that our 
model is well able to
utilize and profit from the inference rules. 
In the oracle setting, SR performs much better than GR. This is expected, since oracle rules with access to the full context will deliver more contextually-relevant inferences, while GR rules may diverge more from the story context. However, in the 
realistic NSC task setting (Table \ref{tab:resultnsc}, lines 5,6) GR outperforms SR, which again underlines the generalization capacities of \COINS. 

\textbf{Impact of different inputs for the Generation Step.} 
In Table \ref{tab:inputtypes} we investigate the performance of \COINS\ with different inputs to
the sentence generation component \textit{at inference time}: 
(i) 
When only inference rules 
(from the inference step) are given 
to the model without any story context ($S'$ = $\{s_1, s_2,$[SEP]$, s_n\}$) (\textbf{IR only}), 
sentence generation 
benefits when specific rules are used. This is expected since
the specific rules contain 
statements with concrete character names and paraphrased events from the story. 
(ii) When only the story beginning ($s_{1,2}$) is provided to the sentence generation model \textit{without} the ending sentence  $s_n$ (\textbf{w/oSE}) nor inference rules (\textbf{w/oIR})
we observe that the performance drops compared to models given the full incomplete context ($S'$),
indicating that knowing the story ending helps the model to generate missing sentences that are
coherent with the 
story. 
However, (iii) when adding inference rules \textbf{IR} (from the inference step i.e., E$_i$ + C$_i$) to the 
context ($s_{1,2}$) 
without ending sentence (\textbf{w/oSE}), performance again improves (+$5.85$ \bleu\ scores). Note that the inference rule contains the \textsc{Cause} relation for $s_n$.
This indicates that the model is able to utilize 
inference rules
for story generation.\footnote{Here, we report the results with generalized rules as GR works better than SR when context is given (cf.\ Table. \ref{tab:resultnsc}).}

\begin{table}[t!]
\centering
{
\scalebox{0.75}{
\begin{tabular}{@{}l@{~~~}c@{~~~}c@{~~~}c@{~~~}c@{~~~}}
\toprule
{\bf Input} & \textbf{PPL} ($\downarrow$) & \textbf{BLEU-1/2} ($\uparrow$) & \textbf{ROUGE-L} ($\uparrow$)\\\midrule
{\bf IR only} (GR)  & 13.05 & 10.65/4.01& 6.31\\
{\bf IR  only} (SR)  & 8.01 & 15.65/6.08& 15.31\\
\bf{No IR + w/oSE} & 11.5 & 15.12/5.95 & 12.47 \\
{\bf IR (GR) + w/oSE} & 7.49 & 21.50/9.78 & 18.07\\
\bottomrule
\end{tabular}}}
\caption{Impact of different inputs to \COINS\ for Story Completion, SR: specific rules, GR: general rules, IR: inference rules, \textbf{w/oSE}: w/o the story ending ($s_n$).} \label{tab:inputtypes}
\end{table}

\paragraph{Performance of inference rule generation.} 
We now investigate how difficult it is to generate contextualized inference rules (specific and general) when multiple sentences are missing from a story. For this we compare \COINS\ to a GPT-2 model fine-tuned on \GLUCOSE\ data to generate inference rules (cf.\ $\S$4).
We study
the impact of jointly and dynamically learning 
sentence and inference rule generation (in \COINS) on the inference generation task -- while 
the fine-tuned GPT-2 model only learns to generate inference rules conditioned on the static story context. 
We
specifically examine
the difficulty of generating inference rules \textit{for two consecutive sentences} ($s_3$ and $s_4$) in a 5-sentence context, as opposed to shorter sequences, in three different scenarios: i) when the \textit{complete story context $S$} is given; ii) when \textit{the incomplete context $S'$} (i.e., $s_1, s_2$ and $s_5$) is given, plus either $s_3$ or $s_4$ (\textbf{1-missing sentence}), and iii) when $S'$ is given, but neither of the intermediate sentences $s_3$ and $s_4$ (\textbf{2-missing sentences}). In each setting, we generate \textsc{Effect} and \textsc{Cause} rules for the targeted sentences $s_3$, $s_4$, and compare their quality. The results are reported in Table \ref{tab:automaticevalknowledge}.
We observe that in the \textbf{2-missing sentences} setting, \COINS\ outperforms 
GPT-2 (by $+2.3$ \bleu\ score on average).
This indicates that 
learning to perform inference rule generation jointly with sentence generation is beneficial for filling-in multiple  story sentences. Interestingly, for increasing numbers of missing sentences, performance drops drastically for \textsc{Cause} (as opposed to \textsc{Effect}), but less so for \COINS\ as opposed to GPT-2.
A 
possible reason for this may be 
the conditional, uni-directional nature of the underlying 
\textsc{GPT-2} language model, which is trained to predict follow-up words in forward direction. This may favor future-directed \textsc{Effect} rules -- as opposed to  \textsc{Cause} relations. 
The milder effect on \COINS\ could indicate that the concurrent inference model supports the sentence generation model to overcome this weakness.\footnote{In future work, we will test the above hypothesis by experimenting with a bi-directional transformer generation model.}
\begin{table}[t!]
\small
\centering
{
\scalebox{0.82}{
\begin{tabular}{@{}l@{~~~}c@{~~~}c@{~~~}c@{~~~}c@{~~~}c@{~~}c@{~}}
\toprule
{}&\multicolumn{2}{c}{Full Context} & \multicolumn{2}{c}{1-Missing Sentence}  & \multicolumn{2}{c}{2-Missing Sentence}\\
{\bf Model} & \textbf{E} &\textbf{C} & \textbf{E} & \textbf{C} & \textbf{E} & \textbf{C} \\\midrule
{\bf GPT-2$^\dagger$} & 58.3& \textbf{63.3} &56.5 & 58.3& 55.4 & 53.9\\
\bf{COINS}&\textbf{59.9}& 62.9 &\textbf{58.6}& \textbf{60.3}&\textbf{57.5}& \textbf{56.8}\\
\rowcolor{gray!30}
{\bf GPT-2$^\dagger$}&57.7&59.5& 55.5& 55.3& 53.4&51.4\\
\rowcolor{gray!30}
\bf{COINS}&57.8& \textbf{60.1}&\textbf{56.3}&\textbf{58.2}&\textbf{55.1}&\textbf{55.2}\\
\bottomrule
\end{tabular}}}
\caption{Automatic evaluation of the quality of inference rules in different context settings. Best results in \textbf{bold}. Metric: \bleu-1 scores, \textbf{E}: \textsc{Effect}, \textbf{C}: \textsc{Cause}, Grey: context-specific rules (SR); regular: general rules (GR), $^\dagger$: \textit{fine-tuned} on \GLUCOSE\ dataset.}\label{tab:automaticevalknowledge}
\end{table}
\begin{table*}[t!]
\small
\centering
{
\scalebox{1}{
\begin{tabular}{@{}l@{~~~}c@{~~}c@{~~}c@{~~}c@{~~}c@{~~~}c@{~~}c@{~~}c@{~~}c@{~~~}c@{~~}c@{~~}c@{~~}c@{~~}}
\toprule
{} &\bf{Knowledge}&\multicolumn{2}{c}\textbf{Coherence} & \multicolumn{4}{c}\textbf{Grammaticality}  \\
\bf{Models} & \bf{of Base Model} & {Win($\%$)} & {Tie($\%$)} & {Loss($\%$)} & {$\kappa$} & {Win($\%$)} & {Tie($\%$)} & {Loss($\%$)} & {$\kappa$} &\\\hline 
{\bf COINS} vs {\bf GPT-2}& None &54.7 & 32.0& 13.3& 0.52& 45.7& 41.3& 13.0&0.49\\
{\bf COINS} vs {\bf \textsc{Gluc.}-GPT-2}& \textsc{Glucose} &52.0& 33.0& 15.0& 0.43& 31.7& 54.3&14.0&0.45\\
{\bf COINS} vs \bf{KE} & \textsc{Cn} + \textsc{Atomic}& 50.0 & 32.0& 18.0& 0.44& 21.3 & 69.7 &9.0&0.37\\
{\bf COINS} vs \bf{GRF} & \textsc{Cn}& 50.5 & 30.5& 19.0& 0.48& 20.5 & 70.0 & 9.5&0.35\\
\bottomrule
\end{tabular}}}
\caption{Manual evaluation 
of sentence generation quality of \COINS\ (GR) 
for 100 
stories. Scores 
are percentages of \textit{Win, Loss,} or \textit{Tie} when comparing \COINS\ to baselines. 
Fleiss’ kappa $\kappa$: 
fair agreement or moderate
agreement. 
}\label{tab:manualeval}
\end{table*}

\if false 
\begin{table*}[t!]
\small
\centering
{
\scalebox{1.0}{
\begin{tabular}{@{}l@{~~~}c@{~~~}c@{~~~}c@{~~~}c@{~~~}c@{~~~}c@{~~~}}
\toprule
{\bf Model} & \textbf{PPL} ($\downarrow$) & \textbf{BLEU-1/2/4} ($\uparrow$) & \textbf{METEOR} ($\uparrow$) & \textbf{ROUGE-L} ($\uparrow$) & \textbf{CIDEr} ($\uparrow$)\\\midrule
{\bf GPT-2} (124M) & 11.56 & 11.79/4.6/1.06 & 12.15 &  12.0 & 14.40\\
{\bf GPT-2} (774M)&  10.22 & 12.62/5.24/1.26 & 13.03 & 12.76 & 17.05\\ 
\bf{Knowledge Enhanced} \cite{guan-etal-2020-knowledge} & &  & \\
\bf{GRF} \cite{ji-etal-2020-language} & {12.18} & 15.66/5.92/1.21 & 13.31 &  13.12 & 12.57\\\hline 
{\bf COINS (Contextualized Rules) (effect-cause)} & &  {17.49/6.8/1.39} & & {17.00} & {19.28}\\
{\bf COINS (Generalized Rules) (effect-cause)} & & \textbf{18.05/7.01/1.47} & & \textbf{17.09} & \textbf{20.94}\\ 
{\bf COINS (Contextualized Rules) (cause-effect)}  & &  &  & & & \\
{\bf COINS (Generalized Rules) (cause-effect)}  & &  &  & & & \\\hline 
{\bf COINS Oracle (Contextualized Rules)} & & {30.75/22.76/15.01} & {28.24} & {32.5} & {122.1}\\
{\bf COINS Oracle (Generalized Rules)} & &  &  & & & \\
{\bf{Human}} &&&&\\
\bottomrule
\end{tabular}}}
\caption{Result:  Automatic evaluation results on the Story Completion Task. The best performance is highlighted in \textbf{bold}.}\label{tab:automaticeval}\label{tab:finetune}
\end{table*}
\fi


\if false
\red{begin: to be replaced/rephrased}
We re-train \COINS\ and the fine-tuned GPT-2 on the NSC training data in the following settings:  
(i) given the \textit{full story context ($S$)} and a sentence ($s_2$) from the story both models learn to generate contextualized inference rules (IR) with respect to \textsc{Effect} and given ($s_5$) sentence and $S$ learn to generate contextualized inference rules (IR) with respect to \textsc{Cause},
(ii) when \textit{one sentence ($s_4$)} \textit{is missing from the story}: given the \textit{incomplete story ($S'$)} and a sentence ($s_{3})$, both models learn to generate contextualized IR with \textsc{Effect} relation, and  similarly given a sentence ($s_{5})$ and $S'$ models learn to generate IR for \textsc{Cause}; (iii) \textit{Two missing sentence:} when \textit{two sentences ($s_3, s_4$)} \textit{are missing from the story $S'$}: 
the model learns to generate IR for sentence ($s_{2}$) with \textsc{Effect} relation and for sentence ($s_{5}$) with \textsc{Cause} relation. 
We test our model's performance on 100 stories from the \GLUCOSE\ test set and compare it to
the baseline model GPT-2 (small) for the above different settings (cf.\ Table \ref{tab:automaticevalknowledge}). We observe that as the number of missing sentences increases, the performance of both GPT-2 and \COINS\ decreases. We observe that the performance of \COINS\ is better than GPT-2 ($+2.3$ \bleu\ score on average) when 2-sentences are missing. This indicates that jointly learning to perform inference rule generation jointly with sentence generation is beneficial. 
\red{end}
\fi 


\section{Manual Evaluation}
Automatic metrics can give us some indication of NLG quality, however, these metrics do not necessarily  reflect the  coherence of generated story
sentences. We thus conduct a human evaluation 
focusing on the grammaticality and coherence of the generated 
sentences in their story context. We conduct pairwise comparisons for randomly sampled 100 instances of our best model, i.e., \COINS\ with GR (according to automatic metrics) with four strong baseline models (GPT-2, \GLUCOSE-GPT-2, GRF, KE). 
For each pair of instances (one from \COINS, 
the other from a baseline model),
we present the generated sentences in their 
story context, and
asked three annotators to give a \textit{preference rating} (\textit{win}, \textit{tie}, \textit{lose}) according to the criteria \textit{grammaticality} and \textit{coherence}. For grammaticality, we present each sentence in isolation and ask 
the annotators to rate 
which sentence is more
fluent, readable, and 
compliant with the English standard usage. 
For coherence, we ask the annotators to assess 
which of the two generated sentences are more logically coherent with each other and the story beginning and ending, in terms of causal and temporal dependencies.
We applied
majority voting among the three annotators to obtain final decisions. 
More details about the annotation are given in \textit{Appendix}. 

The human evaluation results are presented in Table \ref{tab:manualeval}.\footnote{We report 
inter-annotator agreement scores calculated with Fless' kappa $\kappa$ \cite{fleiss1971measuring}, calculated for each comparison. We find moderate or fair agreement.} 
The results show
that 
our model produces more coherent and more grammatically correct sentences
compared to all baselines.
This indicates that with support of learned contextualized inference rules based on \GLUCOSE\ knowledge, our model generates more coherent story sentences that are 
causally  and temporally well connected.
 
\begin{figure}[t]
  \centering
    \includegraphics[scale=1.0,height=5cm, width=0.47\textwidth]{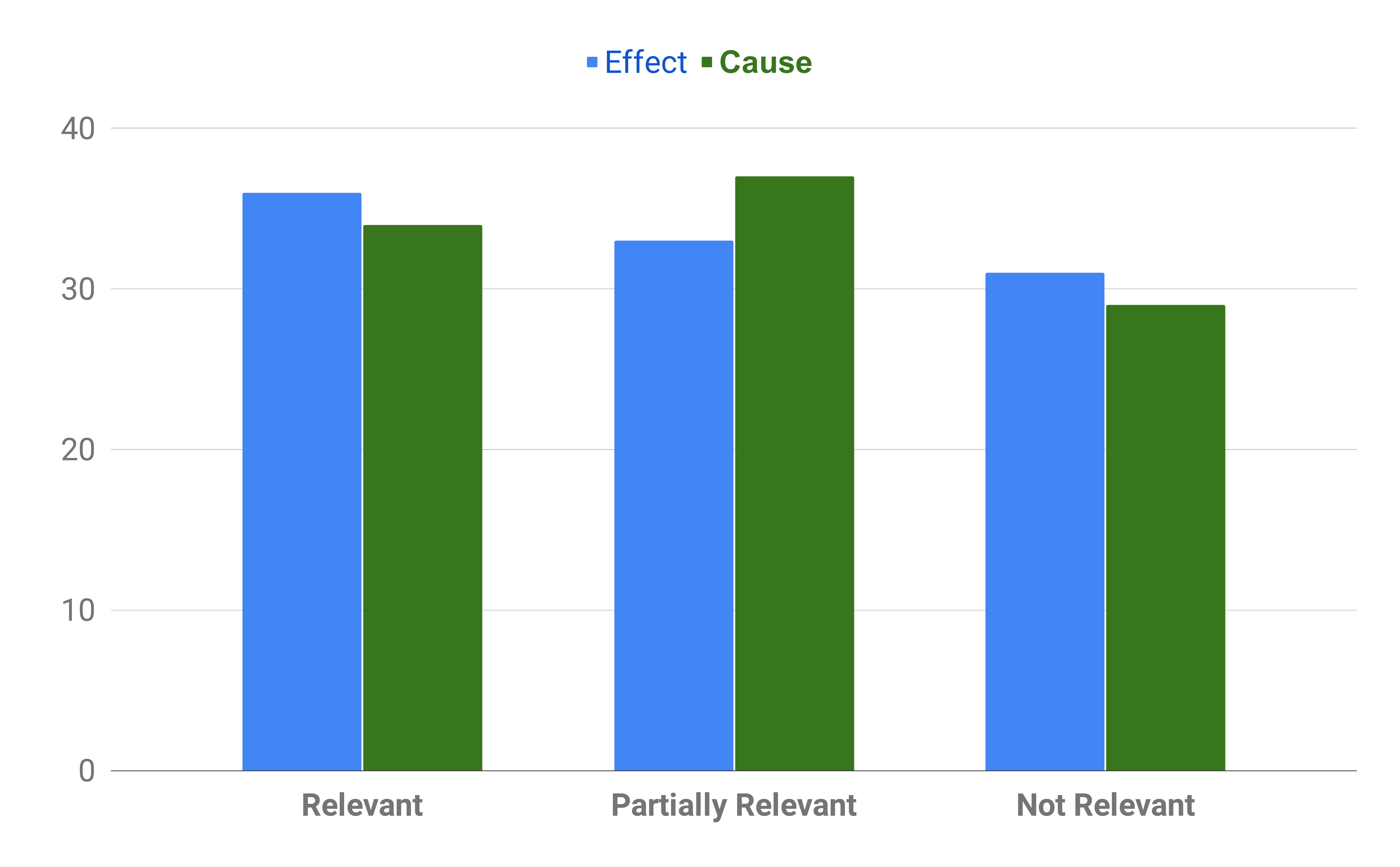}
    \caption{Human evaluation of the relevance of Inference Rules generated by COINS.}
    \label{fig:eval_human_IR}
\end{figure}

\paragraph{Relevance of Generated Inferences Rules.} We further conduct human evaluation to validate the effectiveness and relevance of the generated inference rules. We randomly select 50 instances from the \textsc{NSC} dev set. We asked three annotators to evaluate the (GR) inference rules\footnote{We report only COINS (GR), our best model according to automatic metrics.}. We define an inference rule to be relevant if (a) it 
captures implicit causes and effects of a selected sentence $X$ given an incomplete story $S'$, and (b) it is providing useful explanations for the incomplete story $S'$. The result for this evaluation is shown in Fig.\ref{fig:eval_human_IR}, for \textsc{Effect} and \textsc{Cause} relations. We find
that in 36\% and 34\% of cases for effects and causes, respectively (computed on the basis of majority agreement), our algorithm was able to generate relevant inference rules. Our annotations yielded fair inter-annotator agreement of
Fleiss’ $\kappa = 0.45$.

\paragraph{Case Study.} We provide an example from NSC 
with different generation outputs (Table \ref{tab:examples}). Note that the generated sentences are grounded to the inference rules 
obtained from the inference step. Hence, the rules provide both an intermediate representation and explanations for how knowledge can guide or influence story generation.
We provide more qualitative examples in the Appendix. 

\begin{table}[t]
\centering
\small
\scalebox{0.75}{
\begin{tabular}{@{}p{18mm}|p{75mm}}
\rowcolor{blue!15}
Incomplete Story: & $s_1$: Ken was driving around in the snow. $s_2$: He needed to get home from work. $s_5$: His tires lost traction and he hit a tree. \\
\rowcolor{green!15}
Missing \ \ \ Sentences: & $s_3$: He was driving slowly to avoid accidents. $s_4$: Unfortunately the roads were too slick and Ken lost control. \\
{COINS (I$_{GR}$)} & Someone$_A$ is going Somewhere$_B$ $\succ$Cause/Enables$\succ$ Someone$_A$ is at Somewhere$_B$, Someone$_A$ \orange{\textbf{is driving Something$_A$ fast}} $\succ$Cause/Enables$\succ$ Something$_A$ hits Something$_B$ (that is a tree), Someone$_A$ possess(es) Something$_A$ (that is \blue{\textbf{a car}}) $\succ$Enables$\succ>$ Something$_A$ (tires) lost Something$_B$ (traction)\\
\rowcolor{gray!15}
{COINS (I$_{SR}$)} &  He posses(es) a car $\succ$result in$\succ$ His tires lost traction, He needed to get home $\succ$Enables$\succ$ He drove home, He was \orange{\textbf{driving on ice}} $\succ$ Causes/Enables $\succ$ His tires lost traction, He was driving on ice $\succ$Causes/Enables$\succ$ He \blue{\textbf{lost control}} of his vehicle.  \\
{COINS(MS$_{GR}$)} & He was \orange{\textbf{driving too fast}}. He lost control of his \blue{\textbf{car}}. \\
{COINS(MS$_{SR}$)} & He was \orange{\textbf{driving on ice}}. He \blue{\textbf{lost control of his vehicle}}. \\
{GPT-2} &  He stopped at a gas station. He filled his tank.\\
{GPT-2 GLUCOSE} & When he got to the house he realized he was stuck. Ken had to pull over to get help.\\
{KE} & When he got home, he noticed his tires were flat. He decided to pull over.\\
{GRF} & He pulled over to see what was wrong. He saw that his car was stuck in the snow.\\ 
{Human} & He was going very fast. The street was slippery from the snow.\\\hline
\end{tabular}}

\caption{Examples: inference rules and missing sentences generated by \COINS\ (compared to  \textit{Gold} from \GLUCOSE, Green), as well as baseline model generations.\ Gray:\ \COINS\ (SR);\ Regular:\ \COINS\ (GR);\ MS:\ missing sentences, I:\ inference rules}
\label{tab:examples}
\end{table}
\section{Conclusion}

\if false
We propose a novel setting for Narrative Story Completion task, to test model's ability to complete a narrative story given its beginning and ending. We present a model \textsc{Coins} that iterative learns to generate commonsense inference rules \blue{grounded in the context} and to generate coherent story generation, using the generated inferences as a guide. Human and automatic evaluations show that our best model \blue{outperforms strong commonsense knowledge based generation models}. The recursive nature of our inference-driven \textsc{NSC} model holds potential for \blue{knowledge-driven control in the generation of longer sequences.}

As future work, 
1. It would be very interesting to explore different methods to incorporate the generated inference rules in the sentence generation model. \blue{you have done that ;-)} \\
2. Exploring the relationship between inference rules and discourse in story generation.  \blue{yeah!}\\
3. Exploring the relationship between inference rules and factual knowledge in generation. \blue{difficult!} \\
\fi 

We addressed a Narrative Story Completion task that allows us to probe the coherence  capabilities of a neural generation model. We proposed \COINS, a model that iteratively generates commonsense inference rules grounded in the context and  generates story sentences, using the generated inferences as a guide.  
Human and automatic evaluations show that the model outperforms strong commonsense knowledge-based generation models. 
By individuating the inference rule and sentence generation steps, \COINS\ can make the contribution of commonsense knowledge on story generation transparent. The recursive nature of the inference-driven generation model holds potential for knowledge-driven control in the generation of longer sequences. In future work we will explore how an enhanced memory of generated inferences can realize more complex narrative patterns that diverge from strictly ordered narrative sequences.

\section*{Acknowledgements}

This work has been supported by the German Research Foundation as part of the Research
Training Group “Adaptive Preparation of Information from Heterogeneous Sources” (AIPHES)
under grant No.\ GRK 1994/1. We thank our annotators for their valuable annotations. We also thank NVIDIA Corporation for donating GPUs used in this research.



\bibliographystyle{acl_natbib}
\bibliography{anthology,acl2021}
\appendix

\section{Supplementary}

\subsection{Manual Evaluation.} 
\begin{figure}[t]
  \centering
    \includegraphics[scale=1.0,height=5cm, width=0.47\textwidth]{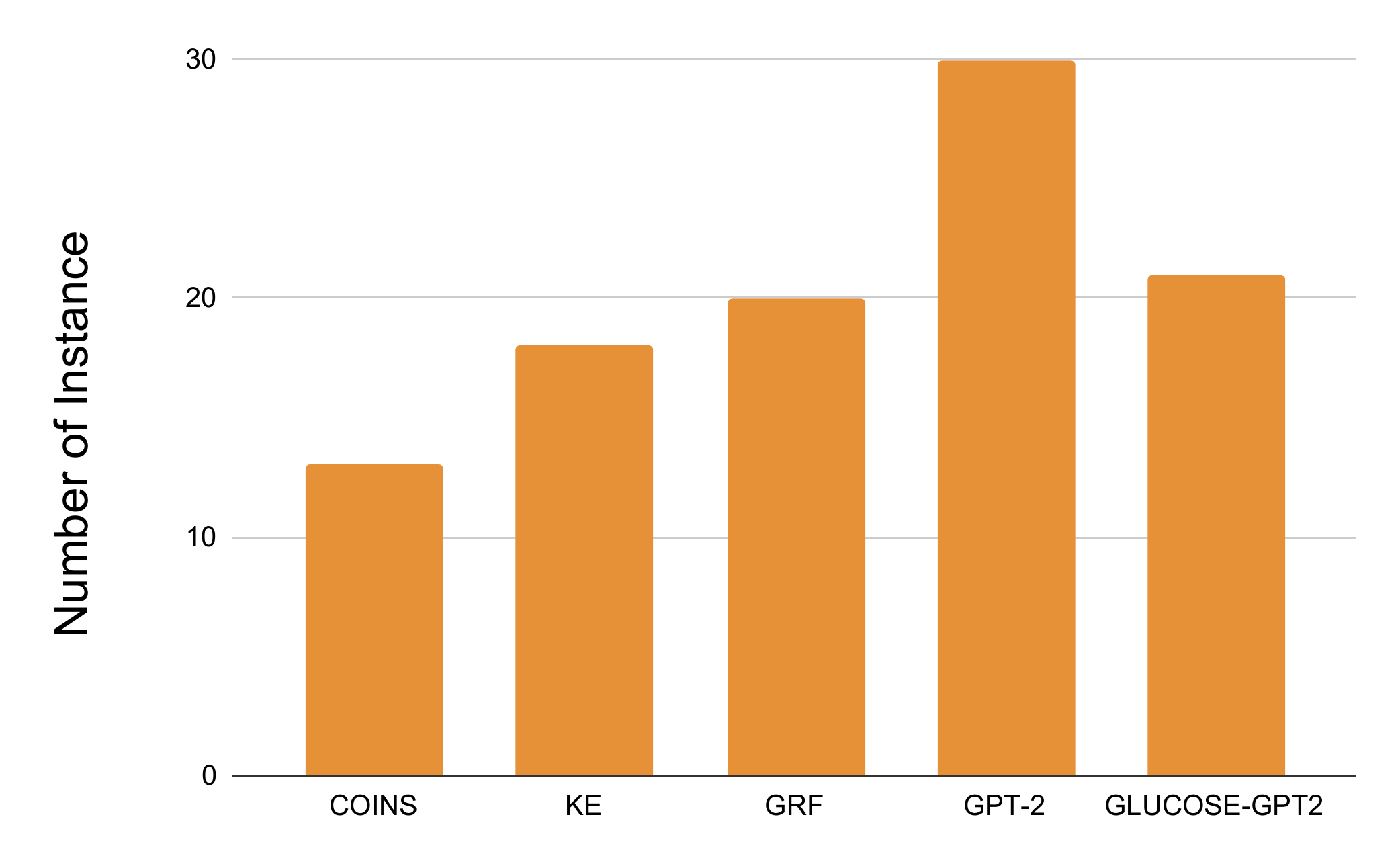}
    \caption{Human evaluation on Contradiction}
    \label{fig:contradiction}
\end{figure}

\begin{figure}[t]
  \centering
    \includegraphics[scale=1.0,height=8cm, width=0.47\textwidth]{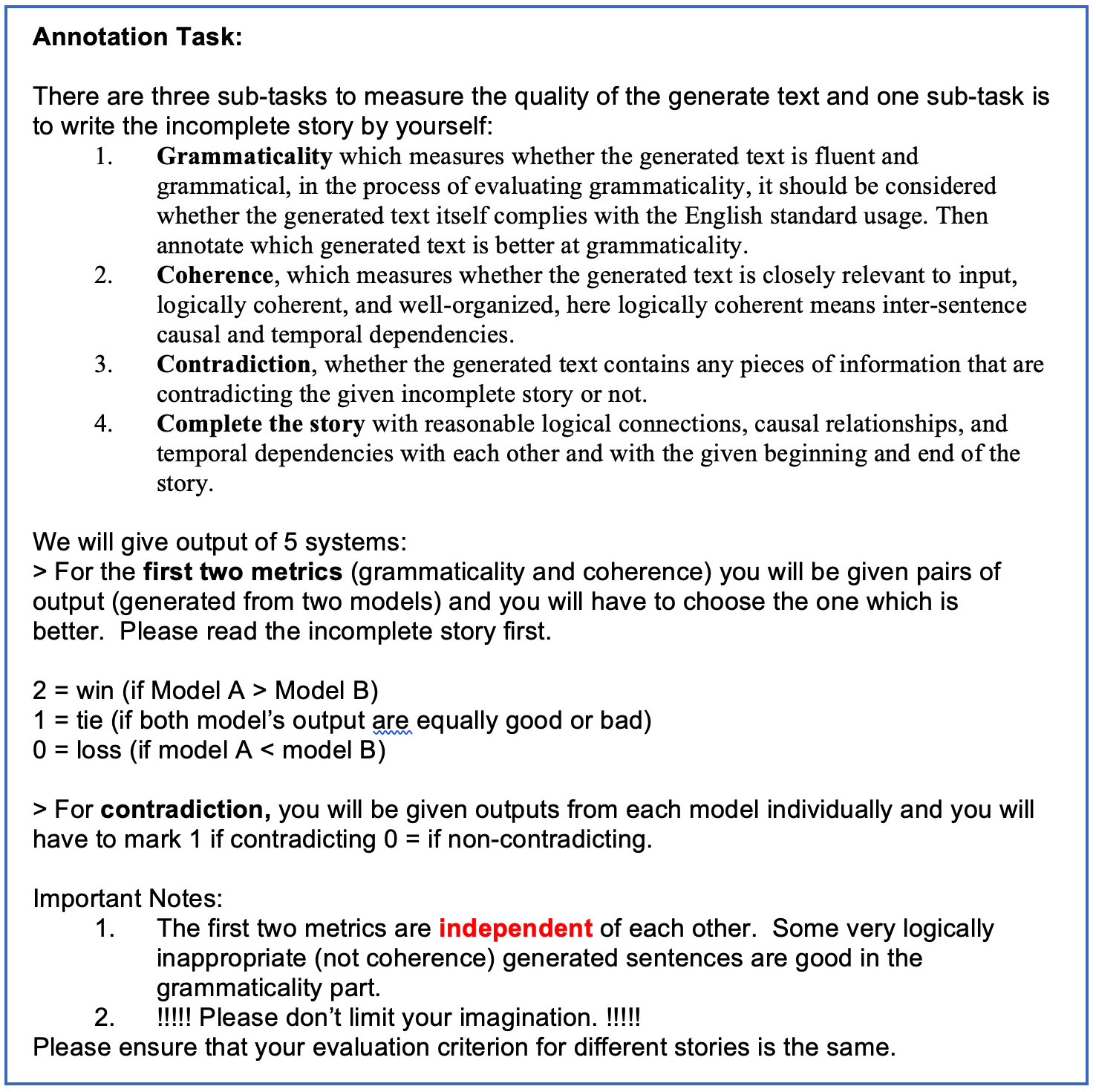}
    \caption{A screenshot of the annotation guidelines for manual evaluation.}
    \label{fig:guide}
\end{figure}

We perform an error analysis to better understand the generation quality. We ask our annotators to assess
whether the generated text contains any pieces of information that are contradicting the given incomplete story or not. Our annotations were performed by three annotators with a \textit{linguistic} background. Figure \ref{fig:guide}, shows a screenshot of the annotation guidelines. Figure \ref{fig:contradiction} depicts the result, we observe the that our COINS models produce less contradicting missing sentences compare to other baselines. 

\subsection{Hyperparameter Details}
\paragraph{Parameter size.} For GPT-2 we use the GPT-2 small checkpoint (117M parameters) based on the implementation of HuggingFace \cite{wolf-etal-2020-transformers} at: \url{https://github.com/huggingface/transformers/tree/master/src/transformers/models/gpt2}\\
\paragraph{Decoding Strategy.} In the inference stage, we adopt beam search decoding with a beam size of 5 for all our models and all baselines we produce. 
We used the following set of hyperparameters for our \textsc{Coins} model: batch size: \{$2, 4$\}; epochs: \{$3, 5$\}; learning rate: \{$1e$-$5$, $5e$-$6$\}. We use Adam Optimizer, and dropout rate = $0.1$. We ran our experiments with
GPU sizes of $11$GB and $24$GB. \\
\paragraph{Training Details.} Our training time is $\approx$24 hours. The original ROCStories Corpus can be found at: \url{https://cs.rochester.edu/nlp/rocstories/}

\subsection{Story Ending Generation Task}
\paragraph{Data.}This task is to generate a reasonable ending given a four-sentence story context \cite{Guan_Wang_Huang_2019}. The stories are from ROCStories \cite{mostafazadeh-etal-2016-corpus}. We use the same  data splits as  \citet{Guan_Wang_Huang_2019}. \\

\begin{table}[t!]
\centering
{
\scalebox{0.7}{
\begin{tabular}{@{}l@{~~~}c@{~~~}c@{~~~}c@{~~~}c@{~~~}}
\toprule
{\bf Model} & \textbf{BLEU-1/2} ($\uparrow$) & \textbf{Distinct-2/3} ($\uparrow$)\\\midrule

{\bf Seq2Seq$^\dagger$} & 19.1 / 5.5 &  0.181 / 0.360\\
\bf{IE+GA$^\dagger$} & 20.8 / 6.4& 0.140 / 0.280\\
\bf{GPT$^\dagger$} & 25.5 / 10.2 & 0.304 / 0.505\\
\bf{GPT2-OMCS$^\dagger$} & 25.5 / 10.4 & 0.352 / 0.589 \\
\bf{GPT2-GLUCOSE} & 25.6 / 10.2 & 0.361 / 0.609 \\
\bf{GRF$^\dagger$} & 26.1 / 11.0 & 0.378 / 0.622\\
{\bf COINS (GR)}  & \textbf{27.4 / 12.3} & \textbf{0.428 / 0.724}\\\hline
{\bf COINS (Oracle)} & 41.80/28.40 & 0.479/0.786\\

\bottomrule
\end{tabular}}}
\caption{Result: Automatic evaluation results on the Story Ending Generation Task, $^\dagger$ \cite{ji-etal-2020-language}}\label{tab:seg}
\end{table}

\paragraph{SEG task.} We also investigate how COINS performs when 
applied to the task of generating a story ending when given a 4-sentence story (SEG).
In this task our model takes only one iteration step to generate the story ending, where in the inference step it generates \textsc{Effect} inference rules for sentence ($s_4$). As seen in Table \ref{tab:seg}, the COINS model outperforms all previous strong baselines, including GPT2-GLUCOSE that uses the same knowledge resource. Interestingly, we also observe that fine-tuning on GLUCOSE or ConceptNet knowledge improves the text generation diversity, indicating that the models leverage concepts and event knowledge during generation (cf.\ Table \ref{tab:seg} line.4-8). \\
\begin{table}[t!]
\small
\centering
{
\scalebox{1}{
\begin{tabular}{@{}llll@{}}
\toprule
{\bf Dataset} & {\bf Train}& {\bf Dev} & {\bf Test}\\
\midrule
\textbf{SEG} & {90,000} &{4,080} & {4,081} \\
\bottomrule
\end{tabular}}}
\caption{Dataset Statistics: nb.\ of unique stories}\label{tab:seg_stat}
\end{table}
\paragraph{Automatic Metrics.} For Story Ending Generation (SEG) we follow the metrics used in \citet{Guan_Wang_Huang_2019, ji-etal-2020-language}:
they use BLEU-1/2 to measure n-gram overlap between  generated and human-written story endings, 
and Distinct-n \cite{li2016diversity} to measure the generation diversity using maximum mutual information. \\
\paragraph{Baselines.} For the \textit{Story Ending Generation task}, we compare COINS to the \textbf{IE+GA} model \cite{Guan_Wang_Huang_2019}. It is based on incremental encoding and multi-source graph attention \cite{Guan_Wang_Huang_2019}.
We also compare to a Seq2Seq model \cite{luong2015effective} based on gated recurrent units (GRU)
and attention mechanism.

\begin{table}[t]
\centering
\small
\scalebox{0.7}{
\begin{tabular}{@{}p{18mm}|p{83mm}}
\rowcolor{blue!15}
Incomplete Story: & $s_1$: Danielle dreamed of living in California. $s_2$: After college she had to decide where to live. [mask] $s_5$: She loved it there. \\
\rowcolor{green!15}
Missing \ \ \  Sentences: & $s_3$: She could move back home or move to California. $s_4$: Danielle decided to take a leap and move to California. \\
{COINS (I$_{GR}$)} & Someone$_A$ decide Something$_A$ (where to live) $>$Causes/Enables$>$ Someone$_A$ decides to live in Somewhere$_A$. \\
{COINS (I$_{SR}$)} & She had to decide where to live $>$Causes/Enables$>$ She chose to \orange{\textbf{live in California}}. \\
{COINS(MS$_{GR}$)} & She decided to live in California. She settled in California. \\
{COINS(MS$_{SR}$)} & She decided to \orange{\textbf{live in California}}. She went to the beach. \\
{GPT-2} &  She finally settled in California. She loved it there.\\
{GPT-2 GLUCOSE} & She decided to move to NH. She found a nice apartment there.\\
{KE} & When he got home, he noticed his tires were flat. He decided to pull over.\\
{GRF} & She decided to move to California. She found a great place to live.\\\hline

\end{tabular}}
\caption{Example1: Generated Inference rules and Missing Sentences}
\label{tab:examples_1}
\end{table}

\begin{table}[t]
\centering
\small
\scalebox{0.7}{
\begin{tabular}{@{}p{18mm}|p{83mm}}
\rowcolor{blue!15}
Incomplete Story: & $s_1$: Her favorite glasses were ruined. $s_2$: The pink dye had gotten all over them. $s_5$: She chose pink, and they both laughed at the irony. \\
\rowcolor{green!15}
Missing \ \ \  Sentences: & $s_3$: Her mother took her to get a new prescription. $s_4$: It was time to order a new pair. \\
{COINS(MS$_{GR}$)} & She took her friend to get a new one. She took it and it was pink.  \\
{GPT-2} &  She bought a new pair of glasses. She wore them to school.\\
{GPT-2 GLUCOSE} & She couldn't decide between two colors. She finally decided on pink.\\
{KE} & She was sad that she couldn't see anymore. Her boyfriend came over to help.\\
{GRF} & She decided to dye them pink instead. She went to the store and bought a pink one.\\\hline

\end{tabular}}
\caption{Example2: Generated Missing Sentences}
\label{tab:examples_2}
\end{table}

\begin{table}[t]
\centering
\small
\scalebox{0.7}{
\begin{tabular}{@{}p{18mm}|p{83mm}}
\rowcolor{blue!15}

Incomplete Story: & $s_1$: Susy was writing an essay by hand for class. $s_2$: She handed it in and thought she would do well. $s_5$: the teacher could not even grade it. \\
\rowcolor{green!15}
Missing \ \ \  Sentences: & $s_3$: But unfortunately the teacher could not even read it. 
 $s_4$: Susy was humiliated. \\
{COINS(MS$_{GR}$)} & But she could not. Teacher didn't read the essay. \\
{GPT-2} &  Suddenly, her hand slipped. She fell and broke her wrist.\\
{GPT-2 GLUCOSE} & But all the sudden she got an F. Susy was so embarrassed.\\
{KE} & When she got her paper back she realized she had tylenol. She had written the entire essay by hand.\\
{GRF} & Susy was very nervous about the essay. The teacher told her she was not allowed to write.\\\hline

\end{tabular}}
\caption{Example3: An example where all the models failed to generated coherent sentences}
\label{tab:examples_3}
\end{table}

\begin{table}[t]
\centering
\small
\scalebox{0.8}{
\begin{tabular}{@{}p{12mm}|p{78mm}}
\rowcolor{blue!15}
Incomplete Story: & $s_1$: Seth was at a party with his friends. $s_2$: \textbf{Someone dared a kid to climb on a wall.} $s_5$: He immediately began screaming that his leg was broken. \\
\rowcolor{green!15}
Missing Sentences: &$s_3$: The kid climbed to the top and everyone cheered. $s_4$: Suddenly he slipped and fell to the ground.\\

\textit{Gold}: & Some People$_A$ (who should not be there) start daring a Someone$_C$ to climb a Something$_C$ (without safety gear) $>$Causes/Enables$>$ Someone$_C$ (who should not be there makes it to the top then falls down and Someone$_C$  (who is acting like monkey)). \\
\rowcolor{gray!15}
&{The kids start daring a kid to climb the wall $>$Causes/Enables$>$ He makes it to the top then falls down and breaks his leg.} \\
\textit{Fine-tuned GPT-2}: &Some People$_B$ start daring a Someone$_A$ to climb a Something$_C$ $>$Causes/Enables$>$ Someone$_A$ quickly shouted that his leg was broken.\\
\rowcolor{gray!15}
{} & Someone start daring a kid to climb the wall $>$Causes/Enables$>$ He shouted that his leg was broken.\\
\textit{COINS}: &  Some People$_B$ start daring a Someone$_A$ to climb a Something$_C$ $>$Causes/Enables$>$ Someone$_A$ is on top of Somewhere$_A$ \\
\rowcolor{gray!15}
 & Someone start daring a kid to climb the wall $>$Causes/Enables$>$ He climbed at the top. \\
\end{tabular}}

\caption{Example of inference rules generated by \COINS\ and \textit{Fine-tuned} GPT-2 when \textbf{2-sentences} are missing (compared to \textit{Gold} from \GLUCOSE). Grey: context-specific rules (SR); regular: general rules (GR).
Bolded sentence $s_2$ is $X$, \textsc{Effect} is the relation type $r$.}
\label{tab:inferenceexamples_}
\end{table}

\if false 
Understanding a narrative story requires both the ability to identify inference rules and also how different events in stories are related in discourse (e.g., by causation, contrast, or temporal sequence) \cite{Mani2012ComputationalMO, mihaylov-frank-2019-discourse}. In this study, we only consider explicit discourse markers\footnote{Discourse markers \cite{hobbs1979coherence, schiffrin1987discourse} are lexical terms such as ‘because’ and ‘but’ that indicate a semantic relation between discourse fragments (such as events).} connecting story sentences (cf. Table \ref{tab:discourse}). To explicitly teach models the discourse relation we append a discourse relation label to each generated inference rule. 
For the relation type \textsc{Effect} \textit{if} there is a discourse marker in the next sentence of the story we append the respective discourse relation to the inference rule otherwise append ``\textit{Succession}". For the relation type \textsc{Cause} \textit{if} there is a discourse marker in the previous sentence of the story we append the discourse relation otherwise append ``\textit{Precedence}".

\begin{table}[!tbp]
      \scalebox{0.7}{
      \centering{
      \begin{tabular}{@{}p{20mm}|p{80mm}}
          \hline
            \textbf{Discourse Rel.} & \textbf{Discourse markers} \\ \hline
            {\textbf{Contrast}} & {however, unfortunately, but, alternatively, instead, despite, although, whereas, though,  notwithstanding} \\
            {\textbf{Cause}} & {therefore, because, thus} \\
            {\textbf{Temporal}} & {whenever , after , before , until , when , finally , during , afterwards , meanwhile} \\\hline
      \end{tabular}}}
      \caption{Discourse relations and their mapped markers. \cite{weisman-etal-2012-learning}}
      \label{tab:discourse}
\end{table}

\af{[again, this is difficult to understand: how does all that relate to Table2? You do not clearly say which relation types you are using, and how this relates to the inference rule dimension. Also you should make clear that this is either a second-round fine-tuning procedure (and based on the silver predicted rules), or else it must be a preprocessing step on the training data, where in case there is a discourse marker in the golden data in the next upcoming sentence, i.e., $X_{i+1}$, you inspect Table 2 (which you have to motivate first), and add the corresponding discourse relation label to the rules that fall under Cause or Effect for what you predict from $X_i$. The next question is then: how does Contrast relate to Cause vs.\ Effect? what do you do with Temporal? If the mapping is only triggered by the discourse marker, I would like to see the distribution: which markers are aligned/mapped to Cause, and which to Effect? Note that there are ambiguous discourse markers: since can be temporal or causal; while can be temporal or concession]}

\fi

\end{document}